\documentclass{article}
\usepackage{spconf,amsmath,epsfig}
\usepackage{gensymb}
\usepackage{adjustbox}
\usepackage{multirow}
\usepackage{booktabs,amsmath}
\usepackage{enumitem}
\usepackage[font={small,it}]{caption}
\usepackage{color,array,dcolumn}

\title{Sea surface temperature prediction and reconstruction using patch-level neural network representations}
\name{Said Ouala $^1$, Cedric Herzet $^{1,2}$, Ronan Fablet $^1$\thanks{This work was supported by GERONIMO project (ANR-13-JS03-0002), Labex Cominlabs (grant SEACS), CNES (grant OSTST-MANATEE) and by MESR, FEDER, Région Bretagne, Conseil Général du Finistère, Brest Métropole and Institut Mines Télécom in the framework of the VIGISAT program managed by "Groupement Bretagne Télédétection" (BreTel).}}
\address{$(1)$ IMT Atlantique; Lab-STICC, Brest, France\\
$(2)$ INRIA Bretagne-Atlantique, Fluminance, Rennes, France}
\begin{document}
%
\maketitle
\begin{abstract}
The forecasting and reconstruction of ocean and atmosphere dynamics from satellite observation time series are key challenges. 
While model-driven representations remain the classic approaches, data-driven representations become more and more appealing to benefit from available large-scale observation and simulation datasets.  
In this work we investigate the relevance of recently introduced bilinear residual neural network representations, which mimic numerical integration schemes such as Runge-Kutta, for the forecasting and assimilation of geophysical fields from satellite-derived remote sensing data. As a case-study, we consider satellite-derived Sea Surface Temperature time series off South Africa, which involves intense and complex upper ocean dynamics. Our numerical experiments demonstrate that the proposed patch-level neural-network-based representations outperform other data-driven models, including analog schemes, both in terms of forecasting and missing data interpolation performance with a relative gain up to 50\% for highly dynamic areas.
\end{abstract}
\section{Introduction}
\label{sec:intro}
The forecasting and reconstruction of ocean dynamics are key challenges. Among others, sea surface geophysical parameters, which can be sensed from space, such as sea surface temperature \cite{chelton_global_2005}, sea surface salinity \cite{klemas_remote_2011} and sea surface height \cite{chelton_chapter_2001}, are important drivers and tracers of the oceanic and atmospheric circulation. Sea surface temperature is for instance a critical parameter in the understanding and forecasting of tropical rainfalls and hurricanes \cite{nobre_variations_1996}.


The forecasting and reconstruction of sea surface geophysical tracers typically rely on model-based approaches which explicitly exploit a dynamical model to perform simulations from given ocean states \cite{gordon_simulation_2000}. The selection and parametrization of a dynamical model however remains a complex issue to capture the variabilities of the spatio-temporal dependencies occurring in the ocean \cite{lorenz_atmospheric_1982}. With the ever increasing amount of observation and simulation data, data-driven approaches have emerged as an appealing strategy to identify explicit or implicit dynamical models. One may cite both analog schemes \cite{lguensat_analog_2017} and neural networks \cite{braakmann-folgmann_sea_2017} as relevant examples of efficient data-driven approaches for the forecasting and reconstruction of sea surface dynamics. 



In this work, we investigate neural network representations for dynamical systems. Neural networks are currently the state-of-the-art techniques for a wide range of machine learning issues. We focus on the representation we recently introduced \cite{fablet_bilinear_2017}. This representation makes explicit the relationship between the neural network architecture and the underlying dynamical system. As detailed hereafter, this representation can be regarded as an implementation of a numerical integration scheme of a dynamical model. Overall, the main contributions of this work are three-fold: i) we propose a new patch-level architecture based on a bilinear residual neural network to capture local SST dynamics. This patch-level representation is applied to the SST anomaly assuming that the large-scale (i.e., horizontal scales above 100km) component of the SST is known. We use a patch-level EOF decomposition to encode the spatial variability, ii) we use the proposed neural network representation in a stochastic data assimilation scheme to solve the reconstruction of SST time series from satellite-derived observations with missing data, iii) we demonstrate the relevance of our proposed model in terms of forecasting and interpolation performances for the case-study region off South Africa.






This paper is organized as follows. Section \ref{sec:DDM} describes the proposed local SST anomaly modelisation architecture. Section \ref{sec:EXP} presents the results of the numerical experiments. We further discuss our contributions in Section \ref{sec:Conc}.

\section{Proposed Model}
\label{sec:DDM}
\subsection{Patch-level representation of SST anomaly}

Given a spatio-temporal SST field $X$, we assume we are provided with an estimate of the large scale component $\overline{X}$ (here for horizontal scales above 100km). This large-scale interpolation typically relies on an optimal interpolation using a parametric covariance model \cite{donlon_operational_2012}. 
The SST anomaly $Y$ is then defined according to (\ref{eq:MSDEC}).

\begin{equation}
{Y}= X-\overline{X}
\label{eq:MSDEC}
\end{equation}

Following \cite{fablet_data-driven_2017}, we consider a patch-level representation of this anomaly field. It comes to decompose $Y$ into overlapping $P \times P$ patches, where $P$ is the width and height of the patch. To encode the spatial variability within each patch, we apply an EOF decomposition learned from a training dataset of anomaly patches. Formally, for a given patch $\mathcal{P}_s$ centered on point $s$, it comes to: 
\begin{equation}
Y(\mathcal{P}_s,t)= \sum\limits_{\substack{k=1}}^{N_E}{\alpha_k(s,t)\beta_k}
\label{eq:EOFD}
\end{equation}
with $\beta_k$ the $k^{th}$ EOF basis and $\alpha_k(s,t)$ the corresponding EOF coefficient for patch $P_s$ at time $t$. The projection of the anomaly patch $Y(\mathcal{P}_s,t)$ in the EOF space is the vector of the $N_E$ coefficients $\alpha_k(s,t)$ denoted as $Z(\mathcal{P}_s,t)$.
The projection matrix $\mathcal{B}$ of the EOF state $Z(\mathcal{P}_s,t)$ into the patch state $Y(\mathcal{P}_s,t)$ can be deduced from the equation \ref{eq:EOFD} by concatenating the basis vectors $\beta_k$ to form the EOF transformation matrix. 

\begin{align}
	Z(\mathcal{P}_s,t) = \begin{bmatrix}
           \alpha_1(s,t) \\
           \alpha_2(s,t) \\
           \vdots \\
           \alpha_{N_E}(s,t)
         \end{bmatrix}
         ,
	\mathcal{B} = \begin{bmatrix}
           \beta_1^t \\
           \beta_2^t \\
           \vdots \\
           \beta_{N_E}^t
         \end{bmatrix}         
  \end{align}
Hence, equation (\ref{eq:EOFD}) may be written in a matrix form:
\begin{equation}
Y(\mathcal{P}_s,t)= Z(\mathcal{P}_s,t)\mathcal{B}
\label{eq:EOFDM}
\end{equation}

\subsection{Neural network representation}

\begin{figure}[htb]

\begin{minipage}[b]{1.0\linewidth}
  \centering
\centerline{\epsfig{figure=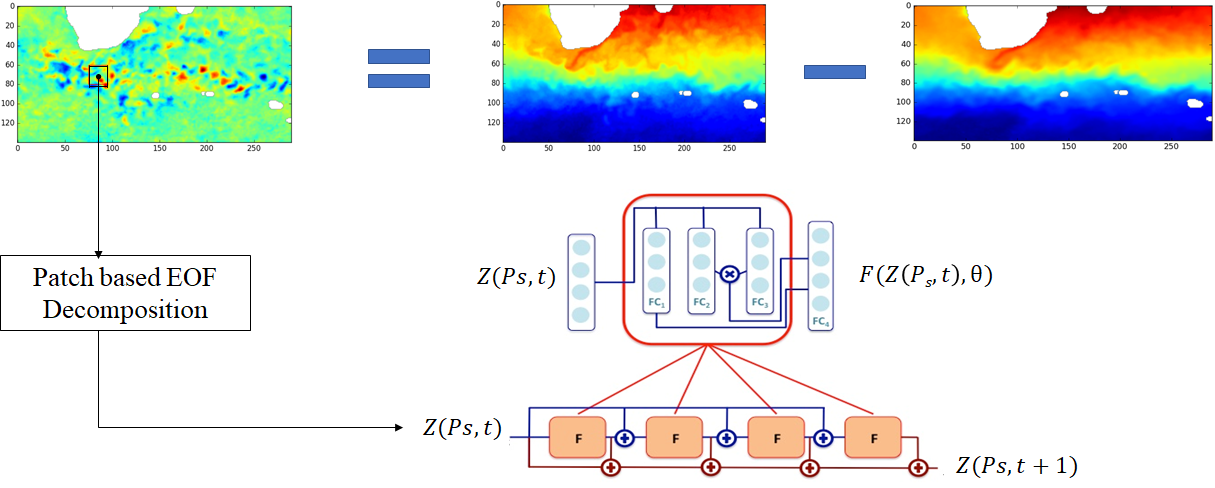,width=8.5cm,height=4.5cm}}
\caption{Sketch of the proposed patch-level neural network representation for SST anomaly: The SST anomaly is computed by subtracting the large scale component to the SST field (top panel) and we apply the proposed neural-network representation to model the dynamics of the SST anomaly at a patch-level. The proposed neural-network architecture mimics a Runge-Kutta-4 numerical integration and a dynamical operator stated as a residual and bilinear neural network \cite{fablet_bilinear_2017}.}
\label{fig:id_sch}
\end{minipage}
\end{figure}

Following our recent study \cite{fablet_bilinear_2017}, we consider bilinear residual neural networks. The key features of these networks are two-fold:
\begin{itemize}[topsep=0pt]
\itemsep0em 
\item They make explicit the relationship between the neural network, the dynamical operator and the associated numerical integration scheme;
\item They embed bilinear terms which are intrinsic features of dynamical systems \cite{brunton_discovering_2016}.
\end{itemize}
Formally, we suppose that the details field is governed by an unknown Ordinary Differential Equation (ODE) in the patch-level EOF space:
\begin{equation}
\frac{dZ(\mathcal{P}_s,t)}{dt}= F\left ( Z(\mathcal{P}_s,t), \theta \right )
\label{eq:sys dyn}
\end{equation}
With $F$ the unknown dynamical operator and $\theta$ the associated parameters. As parameterization for $F$, we consider a bilinear neural network as sketched in Fig.\ref{fig:id_sch}.  

The key aspect of this network is the integration of second order polynomial representations in a fully connected neural network structure. Bilinear layers combine fully connected layers and an element wise product. They are then concatenated to a classical linear layer and feed into an other fully connected layer to compute the residual approximation. The overall representation involves a shared four-blocks architecture, which corresponds to a Runge-Kutta-4 integration scheme for the dynamical operator defined by the above mentioned bilinear residual structure.

This neural network architecture was implemented using Keras framework. For given training data, the learning of model parameters, that is to say the identification of the parameters of the dynamical model, relies on the minimization of the forecasting error using ADAM algorithm \cite{kingma_adam:_2014}.

\subsection{Application to SST forecasting and interpolation}

The learned dynamical model of patch-based EOF-SST anomaly field can be used in several applications. In this work we investigate the relevance of the proposed model for both forecasting and data assimilation issues.

To ensure forecasting, we feed our dynamical model with the first truth EOF decomposition of the SST anomaly. Several forward simulation are then computed using our neural network architecture. We evaluate the forecasting performance in terms of root mean square error (RMSE) for different prediction time steps.

\begin{figure}[htb]

\begin{minipage}[b]{1.0\linewidth}
  \centering
\centerline{\epsfig{figure=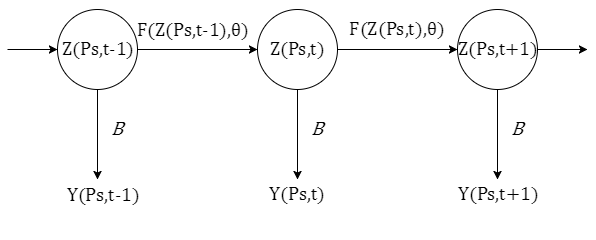,width=8.5cm,height=2.6cm}}
\caption{Stochastic data assimilation process shown as a hidden Markov model : The hidden variables are the patch-level EOF decomposition of the SST anomaly $Z(\mathcal{P}_s,t)$ and the observations are the anomaly patches $Y(\mathcal{P}_s,t)$. We use as dynamical model the proposed patch-level neural network representation trained for location-specific training data. The observation model derives from the EOF decomposition matrix $\mathcal{B}$. We use a classic Ensemble Kalman smoother \cite{evensen_data_2009} solve for the assimilation for each patch.}
\label{fig:HMM_str}
\end{minipage}
\end{figure}

We also consider a data assimilation application as shown in fig. \ref{fig:HMM_str}. Similarly to the analog data assimilation \cite{fablet_data-driven_2017}, the hidden states are considered to be the patch-based EOF decompositions of the SST anomaly and the observations are the detail patches with missing data due to the presence of clouds. The dynamical model is the learned bilinear residual neural network and the observation model is the EOF basis decomposition matrix $\mathcal{B}$. We use a classic Ensemble Kalman smoother \cite{evensen_data_2009} as an assimilation method.

\section{Numerical Experiments}
\label{sec:EXP}
In this section we present the numerical experiments achieved to evaluate the  performance of the proposed framework in terms of forecasting and data assimilation. Our experiments involve a comparison to analog methods \cite{lguensat_analog_2017}.

\subsection{Considered case-study}

The dataset used in our experiments is a gap-free SST time series obtained using the OSTIA product \cite{donlon_operational_2012} delivered by the UK met Office with a $0.05\degree$ spatial resolution from January 2008 to December 2015 with a temporal resolution $h=1$ day. The data from 2008 to 2014 were used as training data and we tested our approach on the 2015 data. The considered region was a region off south Africa located on longitude $5\degree E$ to $75\degree E$ and latitude $25\degree S$ to $55\degree S$.

\begin{figure}[htb]

\begin{minipage}[b]{1.0\linewidth}
  \centering
\centerline{\epsfig{figure=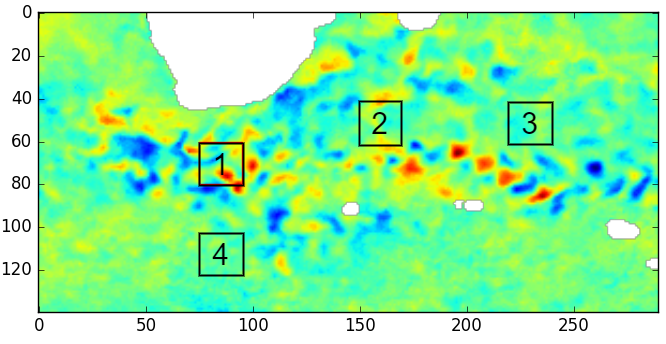,width=8.5cm,height=4cm}}
\caption{Selected patches on the SST anomaly data.}
\label{fig:dsstp}
\end{minipage}
\end{figure}

The patch size used in this work is $P=20$ and the EOF space dimension $N_E=50$, which amounts to capture $95\%$ of the total variance. Figure \ref{fig:dsstp} represents the four patches used to evaluate our method. These four patches were selected to be representative of different dynamic behaviors, patches 1 and 4 being the one which depict the most intense upper ocean dynamics.

\subsection{Experimental setting}
We used a bilinear residual neural network with 4 blocks reproducing the Runge Kutta 4 integration scheme. Two training configurations are considered:
\begin{itemize}[topsep=0pt]
\itemsep0em 
\item Training a single block over the temporal derivative of the anomaly, duplicating the trained block four times to reproduce a Runge Kutta 4 integration scheme. This setting is referred to as Bi-NN(1)-RK4.
\item Training four Bilinear neural network blocks simultaneously over the anomaly time series in a Runge Kutta 4 integration scheme. This setting is referred to as Bi-NN(4)-SL. 
\end{itemize}
It may be stressed that Bi-NN(1)-RK4 and Bi-NN(4)-SL only differ in the way the parameters of the dynamical operator are learnt.
As a bilinear neural network bloc configuration, we used 100 bilinear neurons, 60 linear neurons and 5 fully connected layers with Relu activation functions. 


We compared our results to the recently introduced analog approaches. Analog models have shown great results comparing to the state-of-the-art classical techniques in SST and SSH fields reconstruction \cite{lguensat_data-driven_2017,fablet_data-driven_2017}. The considered analog forecasting operator was based on a locally-linear regression. Two analog operators were used : the local analog forecasting (LAF) and the global analog forecasting (GAF). Regression weights were computed using a Gaussian kernel function. We let the reader refer to \cite{lguensat_data-driven_2017,fablet_data-driven_2017} for additional details on the analog settings.

\subsection{Forecasting and assimilation performance}
We evaluate the forecasting performance at different prediction time steps in terms of RMSE. Results reported in the Tab. \ref{tab:forec1}. Similarly, we report the results of the assimilation experiment in Tab. \ref{tab:assim}. In this assimilation experiment, we consider cloudy SST data using METOP cloud masks from 2015. Both experiments point out the relevance of the proposed neural network representation, which leads to significant relative gain w.r.t. analog methods (up to $\approx$40\% for the assimilation experiment). Interestingly, Bi-NN(4)-SL architecture with shared blocks outperforms Bi-NN(1)-RK4 architecture most of the time. This supports the relevance of truly training the parameters of the dynamical operator within a Runge-Kutta-like setting rather than a simple firs-order approximation as used by Bi-NN(1)-RK4.
 
 


\begin{table}[hbpt]
\caption {{\bf  \em Forecasting performance of data-driven models for the local SST anomaly}: mean RMSE for different forecasting time steps for the following models:  GAF (A), LAF (B), Bi-NN(1)-RK4 (C), Bi-NN(4)-SL (D).}
{\footnotesize
\begin{center}
\begin{tabular}{ll*{6}c}
\toprule
\multicolumn{2}{c}{Model} &A & B & C & D \\
\midrule \midrule 
\multirow{2}{*}{patch 1}
&$t_0+h$  & $1.61$ & $1.58$ & $1.10$ & $\bf1.08$\\
&$t_0+4h$ & $3.70$ & $3.29$ & $3.36$ & $\bf2.76$\\                   
\midrule 
\multirow{2}{*}{patch 2}
&$t_0+h$  & $0.96$ & $0.75$ & $0.36$ & $\bf0.33$ \\
&$t_0+4h$ & $1.35$ & $1.88$ & $1.33$ & $\bf1.11$ \\
\midrule 
\multirow{2}{*}{patch 3}
&$t_0+h$  & $1.09$ & $0.99$ & $\bf0.86$ & $0.87$\\
&$t_0+4h$ & $1.80$ & $1.31$ & $1.32$ & $\bf1.16$\\
\midrule 
\multirow{2}{*}{patch 4}
&$t_0+h$  & $2.34$ & $1.89$ & $\bf1.37$ & $1.44$ \\
&$t_0+4h$ & $1.81$ & $1.55$ & $1.73$ & $\bf1.25$ \\
\bottomrule
\end{tabular}
\end{center} }
\label{tab:forec1}
\end{table}


\begin{table}[hbpt]
\caption{{\bf \em{Assimilation experiment with masked patches observations}}: RMSE of the reconstructed anomaly fields.}
{\footnotesize
\begin{center}
\begin{tabular}{l c c c c }
  \hline
  Model             & Patch1& Patch2& Patch3&patch4\\
  \hline
  Bi-NN(1)-RK       & \bf0.89& 0.44 & 0.59 & \bf0.60\\
  \hline
  Bi-NN(4)-SL       & \bf0.89& \bf0.42 & \bf0.39 & \bf0.60\\
   \hline
  AnDA-G			& 2.10& 1.78 & 3.14  & 0.93\\
  \hline
  AnDA-L			& 0.98& 0.44 & 0.73  & 0.78\\
  \hline
\end{tabular}
\end{center} }
\label{tab:assim}
\end{table}

\section{Conclusion}
\label{sec:Conc}

Overall, through an application to SST anomaly, this study supports the relevance of our recently introduced neural network architectures \cite{fablet_bilinear_2017} for the data-driven prediction and reconstruction of sea surface geophysical fields from partial satellite observations. These neural networks representations outperform for the considered case-study analog methods and provides an explicit interpretation of trained model in terms of dynamical operator. Future work will further explore such neural network representations and their applications to satellite ocean remote sensing data.



\bibliographystyle{IEEEbib}
\bibliography{Zotero}

\end{document}